\documentclass[wcp]{jmlr}

\usepackage{multirow}

\usepackage{xcolor}

\usepackage{xcolor}
\usepackage{ifthen}
\newcommand{\markupdraft}[2]{
    \ifthenelse{\equal{#1}{display}}{#2}{}
    \ifthenelse{\equal{#1}{color}}{\color{#2}}{}
}
\newcommand{\notecolored}[3][]{\markupdraft{display}{{\color{#2}\noindent[Note (#1): #3]}}}
\newcommand{\newcolored}[3][]{{\markupdraft{color}{#2}#3}
    \ifthenelse{\equal{#1}{}}{}{\markupdraft{display}{{\color{yellow!70!black}[#1]}}}}
\newcommand{\del}[2][]{{\markupdraft{display}{{\color{olive}[removed: ``#2''[#1]]}}}} 
\newcommand{\new}[2][]{\newcolored[#1]{red}{#2}}
\newcommand{\nnew}[2][]{\newcolored[#1]{orange}{#2}}
\newcommand{\note}[2][]{\notecolored[#1]{green}{#2}}

\newcommand{\yoi}[1]{\note[Yoichi]{\color{cyan} #1}}
\renewcommand{\del}[2][]{}  
\renewcommand{\markupdraft}[2]{}  

\usepackage{arydshln}

\usepackage{longtable}

\usepackage{booktabs}

\makeatletter
\let\Ginclude@graphics\@org@Ginclude@graphics 
\makeatother

\newcommand{\benchname}{NAS-HPO-Bench-II}

\jmlrvolume{157}
\jmlryear{2021}
\jmlrworkshop{ACML 2021}

\title[{\benchname}]{{\benchname}: A Benchmark Dataset on Joint Optimization of Convolutional Neural Network Architecture and Training Hyperparameters}

\author{\new{\Name{Yoichi Hirose} \Email{hirose-youichi-kc@ynu.jp} \\ 
\Name{Nozomu Yoshinari} \Email{yoshinari-nozomu-ry@ynu.jp} \\ 
  \Name{Shinichi Shirakawa} \Email{shirakawa-shinichi-bg@ynu.ac.jp}\\
  \addr Yokohama National University, Kanagawa, Japan}}

\editors{Vineeth N Balasubramanian and Ivor Tsang}

\begin{document}

\maketitle

\begin{abstract}
The benchmark datasets for neural architecture search (NAS) have been developed to alleviate the computationally expensive evaluation process and ensure a fair comparison. Recent NAS benchmarks only focus on architecture optimization, although the training hyperparameters affect the obtained model performances. Building the benchmark dataset for joint optimization of architecture and training hyperparameters is essential to further NAS research. The existing NAS-HPO-Bench is a benchmark for joint optimization, but it does not consider the network connectivity design as done in modern NAS algorithms. This paper introduces the first benchmark dataset for joint optimization of network connections and training hyperparameters, which we call {\benchname}. We collect the performance data of 4K cell-based convolutional neural network architectures trained on the CIFAR-10 dataset with different learning rate and batch size settings, resulting in the data of 192K configurations. The dataset includes the exact data for 12 epoch training. We further build the surrogate model predicting the accuracies after 200 epoch training to provide the performance data of longer training epoch. By analyzing {\benchname}, we confirm the dependency between architecture and training hyperparameters and the necessity of joint optimization. Finally, we demonstrate the benchmarking of the baseline optimization algorithms using {\benchname}.

 \end{abstract}

\section{Introduction}\label{sec:intro}

While deep learning has succeeded in various practical tasks such as image recognition \citep{alexnet, resnet, densenet} and machine translation \citep{lstm, transformer}, deep neural architectures and learning techniques have many tunable hyperparameters. In such a situation, it is difficult and time-consuming to tune machine learning systems manually. Thus, automated machine learning (AutoML), optimizing components of a machine learning system, is recognized as an important research topic in the machine learning community \citep{automlbook}.

A field of optimizing neural network architectures, called neural architecture search (NAS), is an active research topic in AutoML \citep{nassurvey}. It has succeeded in reducing human efforts to design architectures and has discovered state-of-the-art architectures \citep{efficientnet, mnasnet, evolved}. General NAS methods repeat the following steps: (1) sampling candidate architectures from a search space (candidate architecture pool) based on a specific strategy, (2) training the architectures using training data, and (3) evaluating the trained models using validation data. Because this procedure is the same as naive hyperparameter optimization, we can employ a lot of black-box optimizers, such as evolutionary algorithms \citep{rea,cgp-cnn,regularized_evolution}, reinforcement learning \citep{reinforcementnas, metaqnn}, and Bayesian optimization \citep{nasbot, bananas}, for architecture search. 

Most NAS algorithms have the fixed training hyperparameters, but they affect the performance of the architecture obtained by NAS \citep{naseval}.
Thus, developing joint optimization methods of architecture and training hyperparameters is an important direction in the NAS community to enhance the performance of the obtained architectures.
Most simple way of joint optimization is to apply NAS methods with small modification because joint optimization can be also viewed as black-box optimization.
However, the most joint optimization methods require much computational time due to the repetition of model training, while an exceptional, efficient joint optimization method \citep{autohas} exists.
Also, different methods use different configurations such as search space or data augmentation.
It makes harder to compare joint optimization algorithms under fair conditions.

To tackle the issues and accelerate the research of joint optimization, we produce the first benchmark dataset for joint optimization of convolutional neural network architecure and training hyperparameters.
 Our benchmark dataset, termed {\benchname}, can evaluate joint optimization algorithms under the same conditions at low cost.
The benchmark dataset adopts the product space of modern cell-based architecture search space and space of learning rates and batch sizes as hyperparameter space. The cell-based architecture space contains 4K candidate architectures. Additionally, we examine eight learning rates and six batch sizes as the training hyperparameters. We train the models for 12 epochs three times each using the CIFAR-10 dataset~\citep{cifar} and collect the performance data of 192K configurations. Also, we train 4.8K configurations for 200 epochs and build a surrogate model to predict the performance after 200 epoch training from an architecture and training-hyperparameter configuration. As a result, we provide the exact performance dataset of 192K architectures and training hyperparameters for 12 epoch training and the surrogate model-based performance dataset after 200 epoch training. We also analyze the dataset to understand the relationship between architectures and training hyperparameters.
We show that the best-performed training hyperparameters differ depending on the architecture. The result supports the importance of joint optimization of architecture and training hyperparameters. Finally, we demonstrate the benchmarking of well-known \new{NAS and hyperparameter optimization algorithms} using our proposed dataset.

\begin{figure}[tb]
      \centering
        \includegraphics[width=1.0\linewidth]{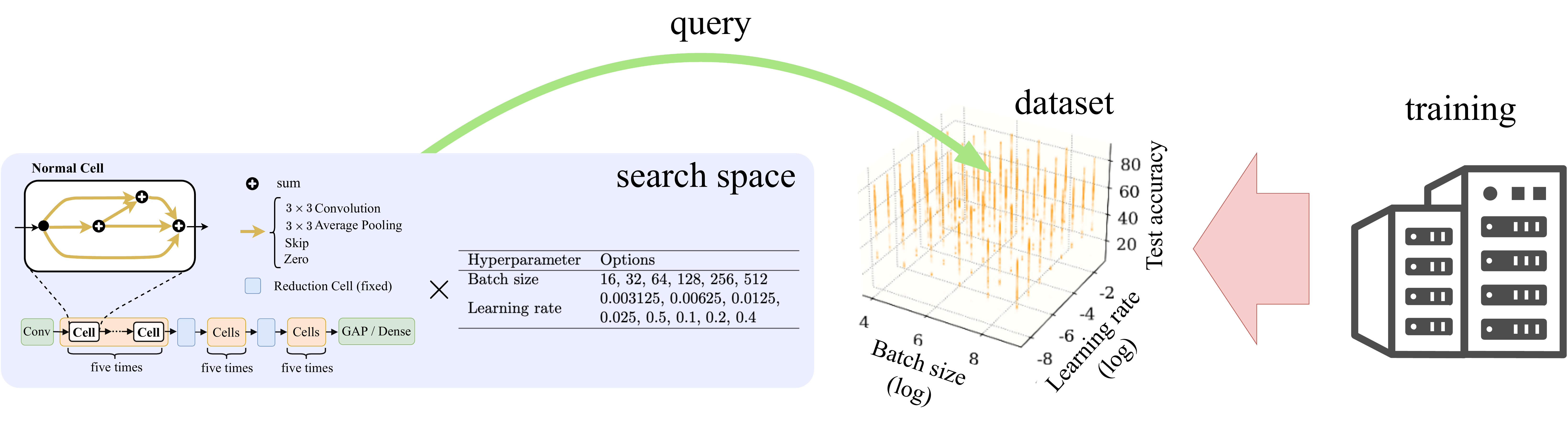}
  \caption{Overview of {\benchname}.}
\label{fig:overview}
\end{figure}

The contribution of this paper is to provide the first benchmark dataset for joint optimization of network connections and training hyperparameters. Figure \ref{fig:overview} illustrates the overview of {\benchname}. Our proposed dataset, {\benchname}, will contribute to accelerating the research on joint optimization of architecture and training hyperparameters.\footnote{The dataset of {\benchname} is publicly \new{available at  \url{https://github.com/yoichii/nashpobench2api}.}}

\section{Related Work}\label{sec:related}
Many benchmark datasets for NAS have been proposed to realize the fair comparison and the reproducibility of NAS algorithms and have reduced the evaluation cost. NAS-Bench-101 \citep{bench101} is the first NAS benchmark dataset consisting of 510M convolutional neural network (CNN) models, in which 423K computationally unique architectures are included. Each model is represented by stacking cells, and each cell is a computational graph whose nodes represent operations such as a convolution operation. NAS algorithms optimize the architecture of the cell. While NAS-Bench-101 has succeeded in the fair comparison of NAS methods, it cannot be used directly to evaluate one-shot NAS methods~\citep{enas, darts, asng-nas}, which greatly reduce the search cost, because the number of operations in the cell varies across the models. The original search space should be modified to evaluate one-shot NAS methods, as done in NAS-Bench-1Shot1~\citep{bench1shot1}.

In contrast, NAS-Bench-201 \citep{bench201} and its extended version NATS-Bench \citep{natsbench} innately support one-shot NAS methods by representing operations as edges of computational graphs, which keeps the number of operations in the cell constant. NAS-Bench-201 trained all 15K CNN models on three image datasets, CIFAR-10, CIFAR-100 \citep{cifar}, and the smaller version of ImageNet \citep{imagenet}, and shows the consistent ranking of model performances across them.
The surrogate dataset NAS-Bench-301 \citep{bench301} provided the model performances in the commonly-used DARTS \citep{darts} search space by predicting performances of $10^{18}$ models from 60K actually trained performance data. Besides, other NAS benchmark datasets have been proposed in addition to CNN and image classification tasks, such as NAS-Bench-ASR \citep{benchasr} for automatic speech recognition, NAS-Bench-NLP \citep{benchnlp} for natural language processing, HW-NAS-Bench \citep{hw-nas-bench} for hardware-aware NAS, and TransNAS-Bench-101 \citep{transnas} for transfer learning.
The NAS benchmarks are leveraged for NAS algorithm evaluation \citep{bananas, alphax}.

NAS-HPO-Bench \citep{nas-hpo-bench} is the only benchmark dataset that includes architecture and training hyperparameters to be optimized. This benchmark dataset permits the comparison of joint optimization methods of architecture and training hyperparameters, but the architecture search space is too simple. Namely, the base model is a multi-layer perceptron (MLP) with two hidden layers, and the architecture hyperparameters are the numbers of units and the types of activations in each layer. Also, the models were trained using the tabular datasets for regression tasks. We note that the architecture search space and target dataset in NAS-HPO-Bench are greatly different from the modern NAS benchmark such as NAS-Bench-101 and 201.
To the best of our knowledge, a NAS benchmark including both layer connectivity and training hyperparameters is not established.

This paper develops a novel benchmark dataset for joint optimization of network connections and training hyperparameters that overcomes the drawbacks of NAS-HPO-Bench. We adopt the larger cell-based CNN architecture search space and image classification task, which is based on modern NAS research. Table~\ref{tab:benchmarks} shows the comparison of existing NAS benchmarks and our proposed benchmark.

\renewcommand*\footnoterule{}

\begin{table}[tb]
\begin{minipage}{\textwidth}
\small
\caption{Comprehensive comparison of NAS benchmark datasets (Size: search space size including equivalent architectures, Arch. type: architecture type to be optimized, Training HPs: presence of training hyperparameters to be optimized, One-shot NAS: support for One-shot NAS, Data: dataset type used for training).
}
\label{tab:benchmarks}
    \centering
    \begin{tabular}{@{}l@{}lc@{}c@{}c@{}c@{}c}
    \hline
        & Name &
         \begin{tabular}{c} Size \end{tabular} &
         \begin{tabular}{c} Arch.\\type \end{tabular} &
         \begin{tabular}{c} Training\\HPs \end{tabular} & 
         \begin{tabular}{c} One-shot\\NAS \end{tabular} &
         Data\\ \hline
         \multirow{8}{*}{\begin{tabular}{c}NAS\end{tabular}}
        & NAS-Bench-101    & 510M & Cell & -- & -- & Image \\ 
        & NAS-Bench-1Shot1 & 363K & Cell & -- & \checkmark & Image \\
        & NAS-Bench-201    & 15K  & Cell &  -- & \checkmark & Image \\ 
        & NATS-Bench       & 32K  & \#channels &  -- & \checkmark & Image \\  
        & NAS-Bench-301\footnote{60K models are trained, and the performance of other models are predicted by a suggorate model.} & $10^{18}$ & Cell & -- & \checkmark   & Image \\ 
        & NAS-Bench-NLP    & 14K  & RNN & -- & -- & Text \\
        & NAS-Bench-ASR    & 13K  & Cell & -- & \checkmark & Audio \\
        & HW-NAS-Bench\footnote{Only the data of latency and energy consumption during inference are provided. Two search spaces are adopted and the data of the larger search space is estimated.} & 15K / $10^{21}$  & Cell &  -- & \checkmark & Image \\
        & TransNAS-Bench-101 & 7K & Cell & -- & \checkmark & Image \\ 
        \hdashline[2.3pt/1.0pt]
        \multirow{2}{*}{\begin{tabular}{c}NAS +\\HPO\end{tabular}}
        & NAS-HPO-Bench      & 62K  & MLP  & \checkmark & -- & Tabular \\ 
        & {\benchname} (Ours)\footnote{Exact performance data of 192K models for 12 epoch training are available. Estimated performance data for 200 epoch training are provided by using a surrogate model.} & 192K & Cell & \checkmark & \checkmark & Image \\ \hline
    \end{tabular}
\end{minipage}
\end{table}
\section{{\benchname}}\label{sec:nashpobench}
This section describes the detail of our benchmark dataset, {\benchname}, for joint optimization of the cell-based CNN architecture, learning rate, and batch size. We use the CIFAR-10 dataset for image classification to train the models because it is a widely-used dataset for NAS evaluation and is also used in existing NAS benchmark datasets. The architecture search space and training configurations are based on NAS-Bench-201 but with some slight differences. We explain the architecture search space design and training hyperparameter design in \ref{subsec:arch} and \ref{subsec:hp}, respectively. We train all models with each training hyperparameters for 12 epochs. Our dataset provides the train/valid/test accuracies, losses, and wall-clock time of all the configurations for every epoch. We further train the randomly selected architectures for 200 epochs to build the surrogate benchmark dataset for 200 epoch training, as done in NAS-Bench-301. We describe the surrogate model construction in \ref{subsec:extra}.

We develop an application programming interface (API) to query performances easily, following existing benchmark datasets. Users can run their NAS methods by querying the training and validation accuracies of 12 epoch for a specific architecture and training hyperparameters and evaluate the discovered architecture and training hyperparameters by querying the surrogate test accuracy of 200 epoch training. Of course, users can use the 12 epoch's test accuracy for NAS method evaluation instead of surrogate data if they do not prefer the surrogate prediction. Using the low-cost API facilitates further developments of joint optimization methods of architecture and training hyperparameters.

\subsection{Architecture Search Space Design}\label{subsec:arch}
The overall CNN architecture is constructed by stacking two types of cells: a normal cell and a reduction cell. The former is architecture-specific, whereas the latter is common in all candidate architectures. The normal cell is represented as a directed acyclic graph (DAG) with four nodes: $G = (\{v_0, v_1, v_2, v_3\}, \{e \mid e=(v_i, v_j) \enspace (0 \le i < j \le 3)\})$. Each edge indicates one of the following operations, and each node indicates the sum of incoming edges.
\begin{itemize}
\item $3\times 3$ convolution (ReLU activation, $3 \times 3$ convolution with stride 1, then batch normalization (BN) \citep{batchnorm}).
\item $3 \times 3$ average pooling with stride 1.
\item Skip, which outputs the input tensor.
\item Zero, which outputs the zero tensor with the same dimension as the input.
\end{itemize}
The zero operation ensures representing an arbitrary depth and width of architectures. Although NAS-Bench-201 includes a $1 \times 1$ convolution, we remove it to reduce the search space size while keeping various operations.
The reduction cell, which halves the image size and doubles the channel size, outputs the sum of two paths.
One path consists of an average pooling with stride 2 and a $3\times3$ convolution.
The other path consists of a $3\times3$ convolution with stride 2 and a $3\times3$ convolution with stride 1.
The model begins with a $3\times 3$ convolution with 16 output channels, followed by batch normalization.
Next, a block of five stacked normal cells is repeated three times connected with the reduction cell.
The model ends with the sequence of batch normalization, ReLU activation, global average pooling, and a fully connected layer.
This architecture search space contains $4^6 = 4096$ possible cell representations.

\begin{figure}[tb]
      \centering
        \includegraphics[width=1.0\linewidth]{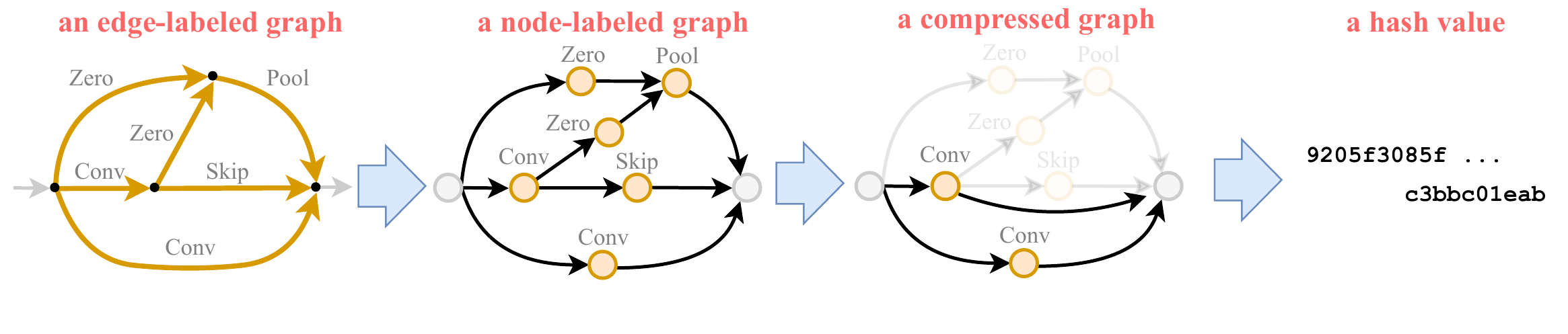}
  \caption{
  The procedure for getting a graph hash value to identify the equivalent cell architectures.
  }
\label{fig:equiv-algo}
\end{figure}

We reduce the architectures to be trained from \nnew{4K to 1K} by identifying the equivalent cell architectures in the search space. First, we convert edge-labeled graphs to node-labeled graphs. Then, we remove the zero operation nodes and their input and output edges and replace the skip operation nodes with edges. We also remove the nodes that are not connected with input or output nodes. Finally, we apply the graph hashing algorithm for node-labeled DAG \citep{dag-hash} to the compressed graphs and get the hash values. We identify the equivalent cell architectures by comparing these hash values. Figure~\ref{fig:equiv-algo} illustrates the procedure for getting a graph hash value.

\subsection{Training Hyperparameter Design}\label{subsec:hp}
We chose the initial learning rate and batch size as the training hyperparameters to be optimized in our benchmark dataset. All the architectures were trained for the combination of eight initial learning rates and six batch sizes, as shown in Table~\ref{tab:lrbatch}. Several settings have the same ratio of batch size to learning rate, which enables detailed analysis with the scaling rule described in Section~\ref{subsec:scaling}.

Other training settings follow NAS-Bench-201.
We use the stochastic gradient descent with Nesterov's momentum of 0.9 for 12 and 200 epochs, the cosine annealing scheduler \citep{sgdr}, and the cross-entropy loss with the weight decay of $5\times 10^{-4}$. 
\begin{table}[b]
\centering
\caption{Training hyperparameter options. The combinations of eight initial learning rates and six batch sizes are used. The learning rate and batch size are set to 0.003125 and 16 as the minimum value, and increase by a factor of 2 to 0.4 and \del{256}\new{512}, respectively.}
\label{tab:lrbatch}
\begin{tabular}{lc}
\hline
Training Hyperparameter & Options \\ \hline
Initial learning rate & 0.003125, 0.00625, 0.0125, 0.025, 0.05, 0.1, 0.2, 0.4 \\
Batch size & 16, 32, 64, 128, 256, 512 \\
\hline
\end{tabular}
\end{table}
We use a single NVIDIA V100 GPU to train each model with each initial learning rate and batch size setting.
The CIFAR-10 dataset was split into a 25K training dataset, a 25K validation dataset, and a 10K test dataset, a typical setting in NAS.
The input pixels are normalized over RGB channels. We apply the random horizontal flip with the probability of 0.5 and the random cropping of a $32 \times 32$ patch with 4 pixels padding on each border as data augmentation.

We train three times with different random seeds in all the configurations. As a result, we get the performance values for 4K architectures and 48 training hyperparameters, a total of 192K combinations, for 12 epoch training. We note that the actual number of model training is 48K.

\subsection{Surrogate Model for 200 Epoch Training}\label{subsec:extra}
We build a surrogate model that predicts the expected test accuracy after 200 epoch training from the architecture and training hyperparameters. To create the dataset for the surrogate model, we train 100 randomly selected architectures three times for 200 epochs using all combinations of initial learning rate and batch size. Therefore, we obtain 4,800 pairs of the hyperparameter and its expected test accuracy as the dataset for the surrogate model.

The surrogate model consists of the bootstrap aggregating (bagging) \citep{bagging} of the 10 base models.
Each base model transforms the normal cell architecture and the one-hot encoded training hyperparameters into two feature vectors using the graph isomorphism network (GIN) \citep{gin} and MLP, respectively. Then, the two feature vectors are concatenated and fed to \new{another} MLP to predict the expected test accuracy.
We perform the architecture-grouped 5-fold cross-validation to evaluate the surrogate model.
In each fold, we keep 20\% of the training data for early stopping \new{as validation data}.
\begin{figure}[tb]
      \centering
        \includegraphics[width=1.0\linewidth]{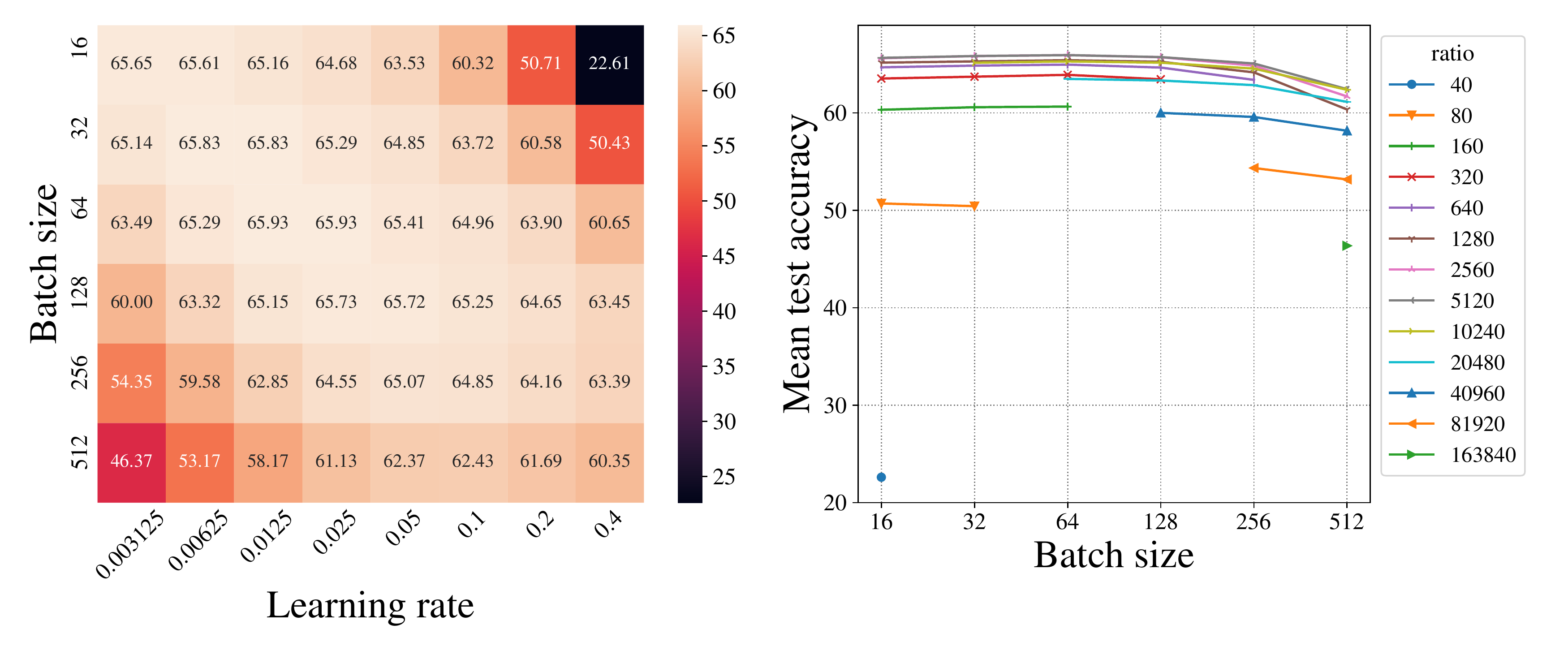}
  \caption{The heatmap on the left shows the mean test accuracy of all the architectures for each training hyperparameter combination.
 The graph on the right is plotted the performance changes of each ratio's group of the scaling rule, showing that the scaling rule also works in our dataset on average.}
\label{fig:overview_heatmap}
\end{figure}
\new{Then, w}e chose the surrogate model at the epoch that recorded the highest $R^2$ score in the validation data and report the test performance. The experimental result shows that our surrogate model achieves a mean $R^2$ score of 0.876 in the test data.
\new{Finally, {\benchname} provides the 200 epoch performance predicted by the surrogate model ensembling five models obtained in the cross-validation process.}

\section{Analysis of {\benchname}}\label{sec:analysis}
This section analyzes our benchmark dataset. We inspect the \new{1K} unique architectures identified by graph hash values to avoid duplication \nnew{and use 12 epoch training results for the analysis.}


\subsection{Data Summary}\label{subsec:summary}
Figure~\ref{fig:overview_heatmap} (left) shows the mean test accuracy of the architectures for each training hyperparameter. We can see that the accuracies vary depending on the combination of hyperparameters. We also observe that the mean test accuracy degrades when using a large batch size and a small learning rate or vice versa. This indicates that a search algorithm should choose better training hyperparameters in our benchmark dataset.

\begin{figure}[tb]
      \centering
  \includegraphics[width=0.7\linewidth]{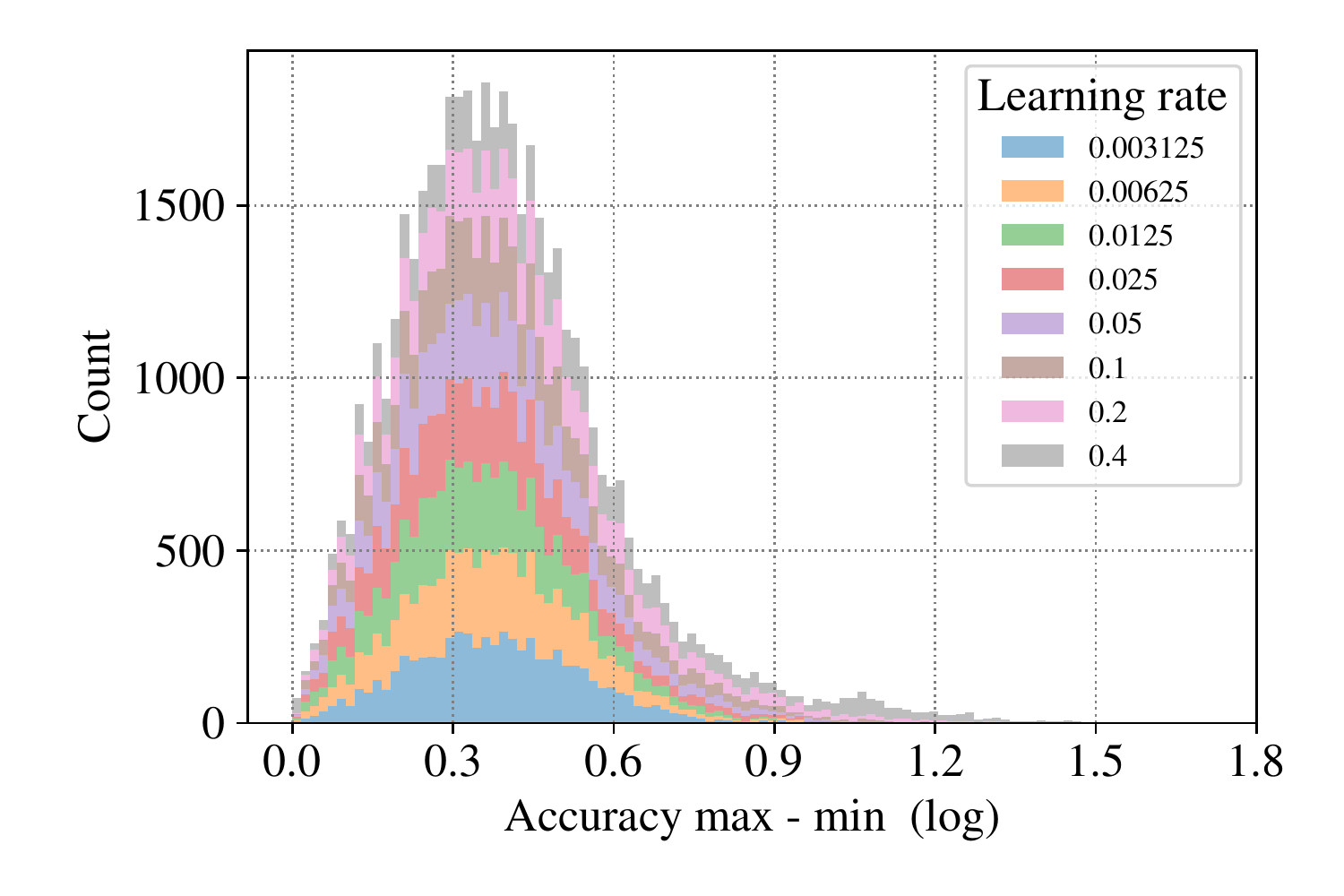}
  \caption{Stacked histogram of the difference between the maximum and minimum test accuracies for \del{10K}\new{1K} unique architectures and 48 training hyperparameters, which is color-coded by learning rates.
  The horizontal axis shows $\log_{10}(1 + d)$, where $d$ indicates the difference between the maximum and minimum test accuracies of three model training.
  }
\label{fig:seed}
\end{figure}

Figure~\ref{fig:seed} shows the histogram of the difference between the maximum and minimum test accuracies among three model training with different seeds. The most frequent value in the histogram is the 1\% accuracy difference, and the values above 1\% account for 63.41\%. The result indicates the randomness for training cannot be ignored. Therefore, we use the mean accuracy of three model training in the later analysis.

\subsection{Scaling Rule}\label{subsec:scaling}
We consider the linear scaling rule \citep{weird-trick, imagenet-one-hour, dont-decay} to visualize and understand the relationship between test accuracy and training hyperparameters.
The linear scaling rule states that the accuracies become similar when the ratio of the batch size to the learning rate is constant as $|\mathcal{B}| / \eta = \new{r}$, where $|\mathcal{B}|$ and $\eta$ denote the mini-batch size and learning rate, respectively, and \new{$r$} is a constant value. We refer to \new{$r$} as the ratio.

We group the settings with the same ratio of \new{$r$}, i.e., such settings are the diagonal elements in the heatmap in Figure~\ref{fig:overview_heatmap} (left). Figure~\ref{fig:overview_heatmap} (right) shows the performance changes of each ratio's group. We observe that the accuracies are stable in the small batch size region, meaning that the scaling rule is also valid in our dataset on average.

\subsection{Dependency of Architecture and Training Hyperparameters}\label{subsec:grouping}
\begin{figure}[tb]
      \centering
  \includegraphics[width=1.0\linewidth]{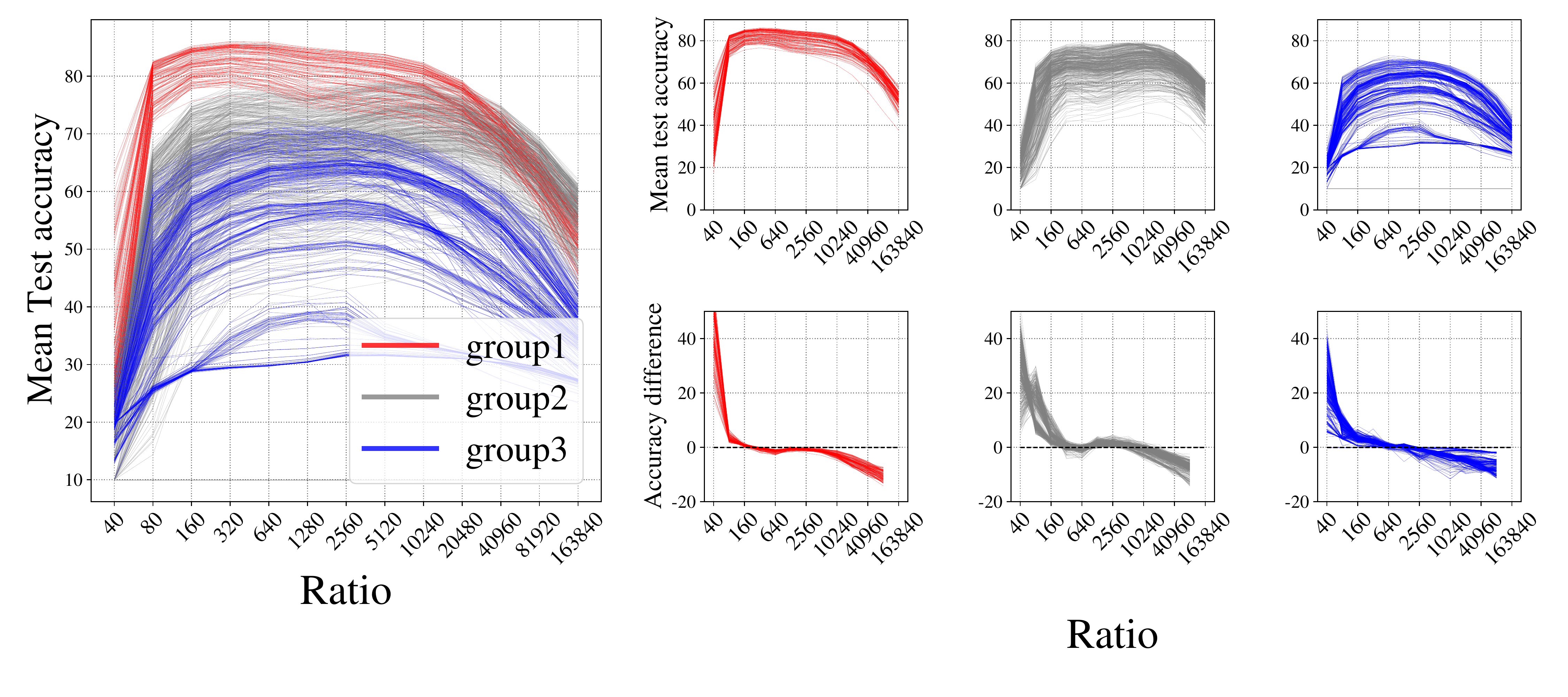}
  \caption{The left side graph shows the relationship between the ratio of the scaling rule and test accuracy.
  Each line indicates the transition of a single architecture. The test accuracies are averaged over the same ratio value of \new{$r$} as $\bar{a}_{(r, m)} = \sum_{h \in H_{r}} f(m, h) / |H_{r}| $, where $m$ represents some architecture, $H_{r}$ is a set of training hyperparameters with the same ratio $r$, and $f$ is a function that outputs test accuracy.
  The upper-right three graphs show the relationship between the ratio and test accuracy of each architecture group. The lower-right three graphs show the transition of accuracy differences of the consecutive ratios, where the vertical axis denotes $\bar{a}_{(2r, m)} - \bar{a}_{(r, i)}$.
  }
\label{fig:group-rel}
\end{figure}

We investigate the dependency of the architecture and training hyperparameters for test accuracy. Figure~\ref{fig:group-rel} (left) depicts the mean test accuracy changes of all \new{1K} unique architectures for the ratio of \new{$r$}. We observe that the best-performed ratio differs depending on the architectures, i.e., the best combination of the learning rate and batch size are different. This result supports the necessity of joint optimization of architecture and training hyperparameters to discover the best-performed configuration.

Next, we deeply analyze the relationship between architecture and training hyperparameters. We can heuristically categorize the architectures into three groups that have a similar trend of accuracy change as follows:
\begin{description}
\item[Group 1] Architectures with one direct edge
\item[Group 2] Architectures not in group 1 and having one or less pooling path and one or more convolutions
\item[Group 3] Architectures not in groups 1 or 2
\end{description}
Here, a direct edge refers to the edge from the input node to the output node in the compressed graph introduced in Section~\ref{subsec:arch}, and a pooling path refers to the path from input to output in which all the nodes are average pooling in the compressed graph.

The colors of the lines in Figure~\ref{fig:group-rel} (left) indicate the architecture groups. Figure~\ref{fig:group-rel} (right) shows how each group's mean test accuracy changes for the ratio of \new{$r$}. Also, the accuracy differences of the consecutive ratios are plotted in Figure~\ref{fig:group-rel} (right).
In group 1, the amounts of change are positive initially and become negative around the ratio value of 320.
In group 2, the amounts of change are positive at the beginning and become near zero around 640, then become positive again and turn to negative around 10240.
In group 3, the amounts of change are positive at first and turn to negative around 1280.
We observe that the training hyperparameters with the ratio around 320 are a good choice for the architectures in group 1, while the ratios around 10240 and 1280 are appropriate for groups 2 and 3, respectively. This result implies that the architecture topology information may be useful to determine the best training hyperparameters.


\section{Benchmarking of Joint Optimization Algorithms}
We evaluate several search algorithms using {\benchname} to demonstrate the use case of our benchmark dataset.
We run well-known NAS and hyperparameter optimization algorithms for 500 trials with a time budget of 20,000 seconds $\approx$ 6 hours: random search (RS) \citep{random-search}, regularized evolution (RE) \citep{regularized_evolution}, REINFORCE \citep{reinforce}, and BOHB \citep{bohb} \new{as joint optimization. Also, we run a combination of BOHB and RE and a combination of RS and RE as sequential optimization (running BOHB/RS first as HPO, then running RE in the remaining time as NAS).} \nnew{Specifically, we initially run HPO for a time budget of 5,000 seconds with a randomly selected architecture and then run the NAS algorithm for the remaining time budget.}
\nnew{The implementation of compared algorithms is based on the publicly available code.\footnote{\url{https://github.com/D-X-Y/AutoDL-Projects/tree/97717d826e1ef9c8ccf4fa29b454946772b2137e/exps/NATS-algos}}} We treat the architecture and training hyperparameters as categorical variables.
In the optimization phase, the validation accuracy of 12 epoch training \nnew{with a specific random seed} is used as the performance measure.
\nnew{The selected architecture and training hyperparameters by the optimization algorithm are evaluated using the test accuracy of 12 epoch training and the predicted mean test accuracy of 200 epoch training provided by the surrogate model.}
{\benchname} allows us to reduce the benchmarking cost of \new{750}\del{11K}\yoi{have calculated wrongly.} GPU days to less than one CPU hour.

\begin{figure}[tb]
  \centering
  \includegraphics[width=0.7\linewidth]{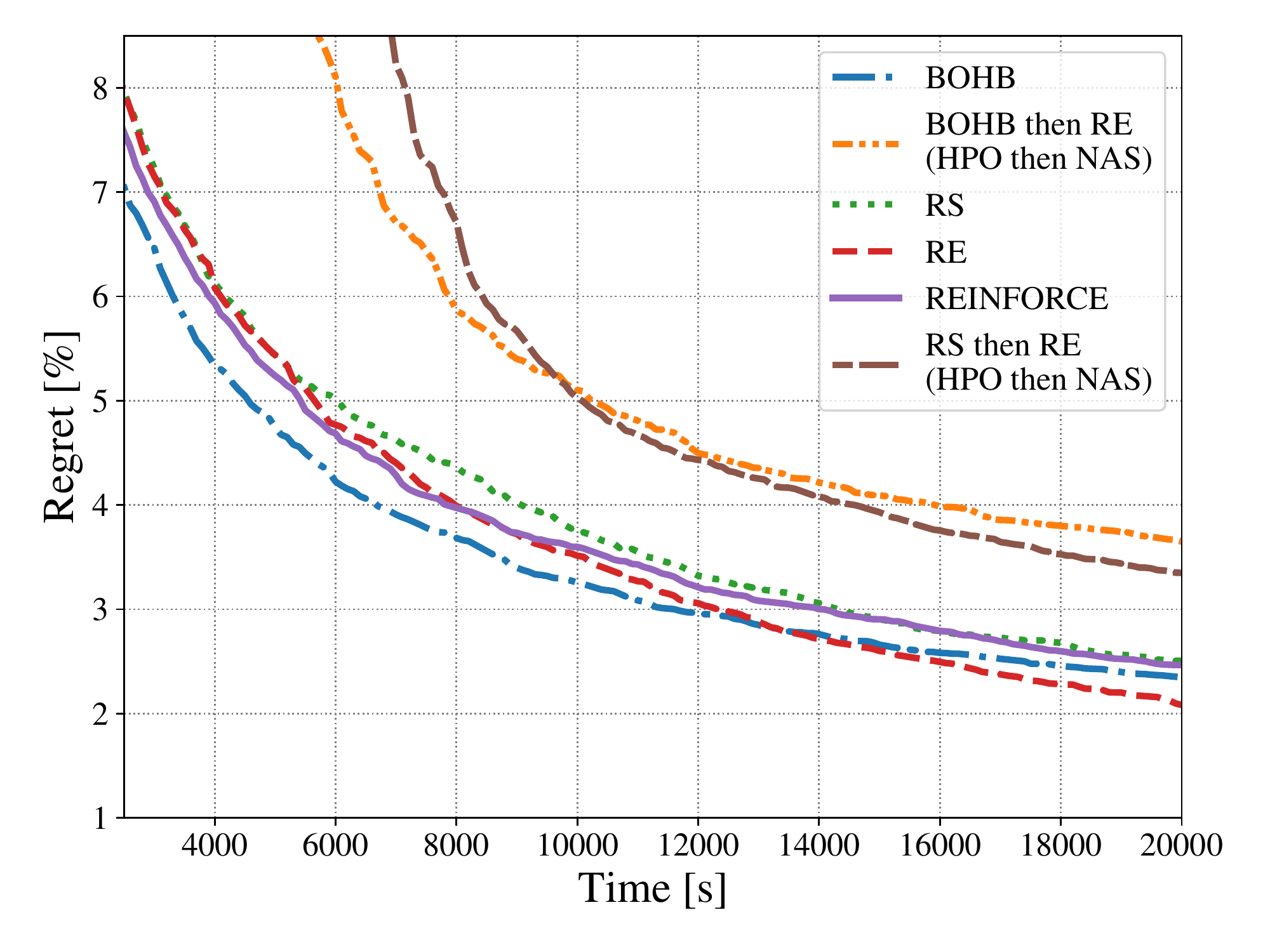}
 \caption{The transition of the mean regret of six algorithms.}
  \label{fig:bench}
  \end{figure}

Figure~\ref{fig:bench} shows the transition of the mean regret of the validation accuracy. The regret is defined by $R_T = f(s^*) - \max_{t \leq T} f(s_t)$, where $s_t$ is the model setting of the $t$-th iteration in a single run, $s^*$ is the best setting in the search space, and $f$ outputs the \del{valid}\new{validation} accuracy of a given model setting.
\new{The result shows joint optimization methods greatly outperform sequential optimization methods in our settings, which suggests the necessity of joint optimization.
Among the joint optimization methods, }regularized evolution reaches the best regret at the end of the optimization, while BOHB shows an early convergence compared to the other algorithms.

Table~\new{\ref{tab:result}} shows the test accuracies of selected model settings by \new{six} algorithms. We report both test accuracies of 12 and 200 epoch training. 
\del{We observe that the rankings of the two types of test accuracies are consistent.}
We have compared the simple baseline algorithms using our dataset. Our dataset may be useful to develop a more sophisticated optimization algorithm for joint optimization of architecture and training hyperparameters.

\begin{table}[tb]
    \centering
    \caption{The test accuracies obtained by \new{six} algorithms. The test accuracies of exact 12 epoch training and surrogate-predicted 200 epoch training are reported. The mean and standard deviation of test accuracies over 500 runs are displayed.}
   \fontsize{8.7}{13}\selectfont
    \begin{tabular}{@{}lc@{\hskip 5px}c@{\hskip 5px}c@{\hskip 5px}c@{\hskip 5px}c@{\hskip 5px}c@{}} \hline
     Algorithm & RS & \new{RE} & REINFORCE & BOHB & BOHB then \new{RE} & RS then \new{RE}\\ \hline
     Result (200 epoch) & $87.41\pm2.14$ & $87.85\pm1.45$ & $87.65 \pm 1.32$ & $87.86 \pm 0.89$ & $87.39 \pm 6.54$ & $87.82 \pm 1.30$\\
     Result (12 epoch) & $83.90\pm1.34$ & $84.32\pm1.71$ & $83.95\pm1.13$ & $84.47\pm 0.92$ & $82.63 \pm 3.84$ & $83.16 \pm 2.26$\\ \hline
    \end{tabular}
    \label{tab:result}
\end{table}

\section{Conclusion and Future Work}\label{sec:conclusion}
We have created {\benchname}, the first benchmark dataset for joint optimization of network connections and hyperparameter optimization in the cell-based CNN. The dataset analysis supports the necessity of joint optimization due to the dependency of architecture and training hyperparameters for test accuracy. We have also demonstrated the comparison of the baseline optimization algorithms using our proposed dataset. We believe that {\benchname} contributes to the further development of NAS and hyperparameter optimization algorithms.

We limited the candidate architecture and training hyperparameters in the search space due to the cost of dataset construction. For example, \new{although the total search space size of our dataset is greater than many existing benchmarks,} the architecture search space is smaller than that of NAS-Bench-201, \new{and we focus on two types of training hyperparameters}. Expanding the search space size is a desired research direction to simulate a realistic scenario\new{, e.g., adding the weight decay coefficient and dropout ratio as training hyperparameters to be optimized}. While this paper focuses on the architecture and training hyperparameters in CNN, it would be worthwhile to create complementary benchmark datasets, such as those treating the data augmentation setting and other deep neural network models.

\acks{\new{We are grateful to Dr. Aaron Klein and Prof. Frank Hutter as the authors of NAS-HPO-Bench for allowing us to call our paper NAS-HPO-Bench-II. This paper is based on results obtained from a project, JPNP18002, commissioned by the New Energy and Industrial Technology Development Organization (NEDO). We used the computational resource of AI Bridging Cloud Infrastructure (ABCI) provided by the National Institute of Advanced Industrial Science and Technology (AIST) for our experiments.}}

\bibliography{acml21}

\end{document}